# Strong completeness and faithfulness in Bayesian networks

Christopher Meek
Department of Philosophy
Carnegie Mellon University
Pittsburgh, PA 15213*

## Abstract

A completeness result for d-separation applied to discrete Bayesian networks is presented and it is shown that in a strong measure-theoretic sense almost all discrete distributions for a given network structure are faithful; i.e. the independence facts true of the distribution are all and only those entailed by the network structure.

## 1 INTRODUCTION

In a series of important papers, Geiger, Verma and Pearl (Geiger et al. 1990 and 1988b and Pearl 1988) outlined an axiomatic approach to characterizing and inferring independence relations in graphical statistical models. A class of graphical models of particular interest are the class of directed acyclic models called Bayesian networks. Pearl introduced d-separation as a rule to infer the independence facts from a given directed acyclic graph; the independence facts are implied to hold of any distribution which factors according to the given directed acyclic graph. An alternative equivalent rule has been proposed by Lauritzen et al. (1990). Geiger et al. (1990) have shown that d-separation is atomic-complete for independence statements for discrete Bayesian networks; one is able to infer all of the independence facts that are logically entailed by the structure of a Bayesian network. In this paper I show that d-separation has the property of strong completeness for discrete Bayesian networks; one is able to infer all of the disjunctive and/or conjunctive combinations of independence statements that are logically entailed by the structure of a Bayesian network. This result shows that d-separation as a rule of inference can not be improved upon even when one restricts attention to discrete variables.

The proof of strong completeness uses a measure-theoretic approach which has important implications for one major approach to learning Bayes networks.

*E-mail address: cm1x@andrew.cmu.edu

Broadly speaking, there are two types of approaches to learning Bayesian networks; the scoring approaches (Bayesian, Likelihood and MDL; see Cooper and Herskovits 1992, Heckerman et al. 1994, Sclove 1994 and Bouckaert 1993) and the independence approaches (see Pearl 1988 and Spirtes et al. 1993). The independence approaches have been shown to be asymptotically reliable assuming that the population distribution stands in a certain relationship to the structure to be learned. The distribution which stands in this relationship to the structure has been called by many names, e.g., faithful (Spirtes et al. 1993), stable (Pearl and Verma 1991) and the structure has been named a perfect map of such a distribution (Pearl 1988). I demonstrate that faithful multinomial distributions exist for every directed acyclic graph and every discrete statespace. Furthermore, in a specific measure-theoretic sense, almost all multinomial distributions are faithful.

The new results in this paper are about discrete Bayesian networks. Strong completeness and the existence of faithful distributions has been shown previously for the class of Gaussian distributions. The discrete (multinomial) case is of special interest since many of the applications of machine learning and data modeling involve discrete data. I include the results for the Gaussian case and give a new and uniform proof of the results for both the Gaussian and multinomial cases.

## 2 STRONG COMPLETENESS AND D-SEPARATION

Pearl introduced d-separation as a rule to infer the independence facts implied by a particular directed acyclic graph. An alternative equivalent rule has been proposed by Lauritzen et al. (1990). Consider the question of whether using d-separation one can infer all of the statements about independence which are entailed by the structure of a Bayesian network. Pearl (Pearl 1988 and Geiger and Pearl 1988b) say that d-separation can not be improved upon. However their proof relies upon Gaussian distributions and thus fails if one restricts attention to the multinomial (discrete) case. In this section it is shown that d-separation is complete even when one restricts attention to multinomial distributions. This bolsters the claim made by Pearl (1988) that d-separation can not be improved upon.[1]

The basic goal of a logic is to derive statements entailed by the assumptions. In the case of the logic of Bayesian networks one is interested in deriving independence statements from a directed acyclic graph $G = \langle \mathbf{V}, \mathbf{E} \rangle$ which are true of any distribution $\mathcal{P}$ from a specific class of distributions over $\mathbf{V}$ for which $G$ is an I-map (see Pearl 1988)[2]. We use $\mathcal{P}$ to denote an arbitrary class of distributions, $\mathcal{P}_\mathcal{N}$ to denote the class of multivariate normal distributions, and $\mathcal{P}_\mathcal{D}$ for the class of multinomial distributions.

Let $i$ range over atomic independence statements, $\mathbf{A} \perp\!\!\!\perp \mathbf{B} | \mathbf{C}$ for disjoint sets $\mathbf{A}$, $\mathbf{B}$ and $\mathbf{C}$. The statement "$\mathbf{A} \perp\!\!\!\perp \mathbf{B} | \mathbf{C}$" is read "$\mathbf{A}$ is independent of $\mathbf{B}$ given $\mathbf{C}$." Let I range over (i) independence statements and (ii) finite conjunctions and disjunctions of I statements. $G$ entails $I$ (written $G \models_\mathcal{P} I$) if and only if $I$ is true in every distribution in $\mathcal{P}$ for which $G$ is an I-map.

As with any logical calculi, there are rules of inference. The central rule of inference in this logic is that of d-separation. The first question one asks about a rule of inference is whether it is sound. The soundness of d-separation as a rule of inference has been demonstrated in Geiger et al. (1988b). The next question one asks about a set of inference rule is whether the set of rules is complete; whether all of the true statements are derivable. A sentence $I$ is derivable by a set of rules $\mathcal{D}$ from assumptions $G$ (written $G \vdash_\mathcal{D} I$) if and only if there is a proof of $I$ from $G$ using the rules of inference $\mathcal{D}$. Geiger et al. (1990) have shown that d-separation as a rule of inference is atomic-complete for the multinomial and multivariate normal class of distributions; in this case $\mathcal{D}$ is simply d-separation.[3]

**Theorem 1 (Atomic completeness; Geiger et al.)** $G \models_{\mathcal{P}_\mathcal{D}} i$ *if and only if* $G \vdash_\mathcal{D} i$.

**Theorem 2 (Geiger et al.)** $G \models_{\mathcal{P}_\mathcal{N}} i$ *if and only if* $G \vdash_\mathcal{D} i$.

Thus, for any given atomic independence fact $i$ one can use d-separation to check if the independence statement $i$ is entailed by the graphical structure. This does not allow us to check independence sentences which are disjunctive combination or sentences of disjunctive and conjunctive combinations of such statements. To derive disjunctive and conjunctive combinations of independence statements the set of inferential rules $\mathcal{D}$, i.e. d-separation, must be expanded to a set of inferential rules $\mathcal{D}*$ which includes $\land$-introduction and $\lor$-introduction.[4]

**Theorem 3 (Strong completeness)** $G \models_{\mathcal{P}_\mathcal{D}} I$ *if and only if* $G \vdash_{\mathcal{D}*} I$.

**Theorem 4 (Geiger et al., Spirtes et al.)** $G \models_{\mathcal{P}_\mathcal{N}} I$ *if and only if* $G \vdash_{\mathcal{D}*} I$.

The proofs of the strong completeness theorems are sketched in the appendix. As with any completeness proof, if a disjunctive independence sentence $\mathbf{A} \perp\!\!\!\perp \mathbf{B} | \mathbf{C} \lor \mathbf{X} \perp\!\!\!\perp \mathbf{Y} | \mathbf{Z}$ is not true for a graph $G$ one must show that there is a model — in our case a probability distribution — in which both $\mathbf{A} \perp\!\!\!\perp \mathbf{B} | \mathbf{C}$ and $\mathbf{X} \perp\!\!\!\perp \mathbf{Y} | \mathbf{Z}$ are not true. Let graph $G$ be given and assume that it is not that case that $G \models_{\mathcal{P}_\mathcal{D}} \mathbf{A} \perp\!\!\!\perp \mathbf{B} | \mathbf{C}$. Geiger et al. (1990) gave a method for constructing a distribution $P$ for which the given graph $G$ is an I-map such $\mathbf{A} \perp\!\!\!\perp \mathbf{B} | \mathbf{C}$ is false in $P$. I extend the result to show that there exists a distribution for arbitrary disjunctive combinations of non-entailed independence facts.

## 3 ASSUMPTIONS FOR RELIABLY LEARNING BAYES NETWORKS

There are several algorithms which use independence tests to learn Bayesian networks from sample data including the PC, and SGS algorithms (Spirtes et al. 1993). Basically these algorithms perform a series of statistical tests of independence using the sample data and based upon the results of these tests the algorithms eliminate a set of possible models until the remaining set of models can not be distinguished by independence facts. The methods enumerated above differ in the series of independence test that are used; the selection and ordering of the tests can improve the practical reliability and computational tractability of these algorithms. Let $S_G$ be any arbitrary boolean combination of independence statements about variables in graph $G$; I write $S$ when the appropriate graph is clear from context. Interpret $\neg \mathbf{X} \perp\!\!\!\perp \mathbf{Y} | \mathbf{Z}$ to mean that $\mathbf{X}$, and $\mathbf{Y}$ are conditionally dependent on $\mathbf{Z}$. A distribution $P$ is **faithful** to the graphical structure $G$ if and only if exactly the independence facts true in $P$ are entailed by the graphical structure $G$. $G$ **faithfully entails** $S$ (written $G \models_\mathcal{P}^\mathcal{F} S$) if and only if $S$ is true in every distribution in $\mathcal{P}$ which is faithful to $G$. It is easy to show that for all directed acyclic graphs $G$ and for all $S$ that $G \models_\mathcal{P}^\mathcal{F} S$ or $G \models_\mathcal{P}^\mathcal{F} \neg S$. Using this fact one can show the theoretical reliability of these algorithms assuming the correctness of the statistical tests and that the population distribution is faithful to the underlying graphical structure. Let $test_i$ be the result of the $i^{th}$ test (e.g. $\mathbf{X} \perp\!\!\!\perp \mathbf{Y} | \mathbf{Z}$ or $\neg \mathbf{X} \perp\!\!\!\perp \mathbf{Y} | \mathbf{Z}$ for disjoint subsets $\mathbf{X},\mathbf{Y},\mathbf{Z}$ of vertices).[5] From the assumption, models can be eliminated in the following way. After performing $t$ tests one can eliminate a model $G$ if $\neg \bigwedge_{i=1}^{t} test_i$ is faithfully entailed by $G$. Eliminate models until the remaining set of models are not distinguishable by conditional independence facts.[6]

---

[1] Pearl's claim is supported by the proof that for any directed acyclic graph $G$ there exists distribution for which $G$ is a perfect map (or, in the language of this paper, the distribution is faithful to $G$); see Theorem 10 of Pearl 1988 and Geiger and Pearl 1988b for a proof. However, the distribution which they construct is a Gaussian distribution. One of the main results of this paper is to show that for any directed acyclic graph and fixed discrete statespace there is distribution to which that graph is perfect map.

[2] Assuming that $P$ is a probability density function then $G$ is an I-map for $P$ if and only if $P$ satisfies the local directed Markov condition with respect to $G$ (see Lauritzen et al. 1990 and Spirtes et al. 1993).

[3] The analogous results for undirected graphs is given as theorem 2.3 in Frydenberg (1990).

[4] The analogous results for undirected graphs is given Geiger and Pearl (1988a).

The assumption of faithfulness has been criticized by several researchers. The essence of the criticism is captured by the following question. How can one ever be confident that the population is faithful to the underlying structure?

This is a reasonable question but an even stronger question seems warranted. Are there faithful distributions (in the class of distributions of interest) for any arbitrary directed acyclic graph? The theorems below demonstrate that the answer to this question is affirmative. The proof of existence for $\mathcal{P}_\mathcal{N}$ uses an alternative proof technique as compared to the proof given in Geiger and Pearl (1988) and Spirtes et al. (1993).

**Theorem 5 (Existence)** *For all directed acyclic graphs $G$ there exists a $P \in \mathcal{P}_\mathcal{D}$ which is faithful to $G$.*

**Theorem 6 (Geiger et al., Spirtes et al.)** *For all directed acyclic graphs $G$ there exists a $P \in \mathcal{P}_\mathcal{N}$ which is faithful to $G$.*

But these theorems do not answer the criticism of the unreasonableness of the assumption of faithfulness. The next theorem shows that at least in a measure-theoretic sense the assumption of faithfulness is reasonable. The distributions in $\mathcal{P}_\mathcal{D}$ and $\mathcal{P}_\mathcal{N}$ are parametric distributions. Let $\pi_G^\mathcal{D}$ be the set of linearly independent parameters needed to parameterize a discrete distribution for which graph $G$ is an I-map and let $\pi_G^\mathcal{N}$ be the set of linearly independent parameters needed to parameterize a multivariate normal distribution for which graph $G$ is an I-map.

**Theorem 7 (Measure zero)** *With respect to the Lebesgue measure over $\pi_G^\mathcal{D}$, the set of distributions which are unfaithful to $G$ is measure zero.*

**Theorem 8 (Measure zero; Spirtes et al.)** *With respect to the Lebesgue measure over $\pi_G^\mathcal{N}$, the set of distributions which are unfaithful to $G$ is measure zero.*

The following interpretation of these results may be helpful. Fix a directed acyclic graph $G$ and a set of parameters; $\pi_G^\mathcal{N}$ in the case of the Gaussian distribution and $\pi_G^\mathcal{D}$ in the case of the discrete distribution. In the case of the discrete distribution the parameters in $\pi_G^\mathcal{D}$ encode all of the possible conditional probability tables for a given directed acyclic graph $G$ with a fixed statespace. Consider any *smooth* distribution over the possible parameter values.[7] Given a distribution over the possible parameterization of the Bayesian network one can consider the probability of drawing a certain type of distribution. The probability of drawing an unfaithful distribution is zero. While this result has clear implications for the existence of faithful distributions and for the strong completeness of d-separation the implication for learning Bayesian networks is less clear; it suggests that the interesting issues about reliably inferring Bayesian networks from data (rather than a population distribution) have to do with near violations of faithfulness.

## 4 FINAL REMARKS

In this section I will give alternative definitions of atomic- and strong completeness to further highlight the distinction between the two concepts, and conclude with a conjecture.

A set of distributions $\mathcal{P}$ is atomic complete for a set of graphs $\mathcal{G}$ if and only if for all graphs $G \in \mathcal{G}$ and for all disjoint sets of vertices $\mathbf{A}$, $\mathbf{B}$, and $\mathbf{C}$ there exists a distribution $P \in \mathcal{P}$ such that $G \models \mathbf{A} \perp\!\!\!\perp \mathbf{B} | \mathbf{C}$ if and only if $\mathbf{A} \perp\!\!\!\perp \mathbf{B} | \mathbf{C}$ is true in $P$. A set of distributions $\mathcal{P}$ is strong complete for a set of graphs $\mathcal{G}$ if and only if for all graphs $G \in \mathcal{G}$ there exists a distribution $P \in \mathcal{P}$ such that for all disjoint sets of vertices $\mathbf{A}$, $\mathbf{B}$, and $\mathbf{C}$ it is the case that $G \models \mathbf{A} \perp\!\!\!\perp \mathbf{B} | \mathbf{C}$ if and only if $\mathbf{A} \perp\!\!\!\perp \mathbf{B} | \mathbf{C}$ is true in $P$. Thus strong completeness differs from atomic completeness in that there is single distribution in which all and only the entailed independence facts hold whereas atomic completeness only requires that some such distribution in the class exists for each non-entailed independence fact.

I conjecture that the proof techniques used for the proofs in this paper can be extended to prove analogous measure zero, existence of a faithful distribution and strong completeness results for the conditional Gaussian class of distributions with respect to directed acyclic graphs (see Whittaker 1990).

---

[5] As above one can define a logical calculus for faithful derivability ($G \vdash_\mathcal{F} S$) using the rule of d-separation to derive both independence and dependence facts. By adding a complete set of propositional inference rules one can show that this logical calculus is strongly-complete for $S$ sentences.

[6] The class of models which are not distinguishable by conditional independence facts constitutes a Markov equivalence class of directed acyclic graphs; see Frydenberg 1990.

[7] The set of distribution for which the claim holds is the class consisting of all measures dominated by the Lebesgue measure.

## Acknowledgements

I would like to thank Clark Glymour, Thomas Richardson, Peter Spirtes and two anonymous referees for helpful comments on an earlier draft of this paper. Research for this paper was supported by the Office of Navel Research grant ONR #N00014-93-1-0568.

## References

Bouckaert, R. (1993). Probabilistic network construction using the minimum description length principle. In *Lecture notes in comp. sci. 747*, pp. 41–48.

Cooper, G. and E. Herskovits (1992). A Bayesian method for the induction of probabilistic networks from data. *Machine Learning 9*.

Dawid, A. (1979). Conditional independence in statistical theory. *J. Roy. Stat. Soc. A 41*(1).

Frydenberg, M. (1990). Marginalization and collapsibility in graphical interaction models. *Annals of Statistics 18*(2).

Geiger, D. and J. Pearl (1988a). Logical and algorithmic properties of conditional independence. Technical Report R-97, Cog. Sys. Lab., UCLA.

Geiger, D. and J. Pearl (1988b). On the logic of influence diagrams. In *Proc. $4^{th}$ workshop on Unc. in AI*, Minneapolis, MN, pp. 136–147.

Geiger, D., T. Verma, and J. Pearl (1990). Identifying independence in Bayesian networks. *Networks 20*.

Heckerman, D., D. Geiger, and D. Chickering (1994). Learning Bayesian networks: The combination of knowledge and statistical data. Technical Report MSR-TR-94-09, Microsoft Research.

Lauritzen, S., A. Dawid, B. Larsen, and H. Leimer (1990). Independence properties of directed Markov fields. *Networks 20*.

Okamoto, M. (1973). Distinctness of the eigenvalues of a quadratic form in a multivariate sample. *Annals of Statistics 1*(4).

Pearl, J. (1988). *Probabilistic Reasoning in Intelligent systems*. San Mateo: Morgan-Kaufmann.

Pearl, J. and T. Verma (1991). A theory of inferred causation. In Allen, Fikes, and Sandwall (Eds.), *Principles of knowledge representation and reasoning: Proc. of the $2^{nd}$ Int. conf.*, pp. 441–452. San Mateo: Morgan Kaufmann.

Sclove, S. (1994). Small-sample and large-sample statistical model selection criteria. In *Selecting Models from Data*, pp. 31–41. Springer-Verlag.

Spirtes, P., C. Glymour, and R. Scheines (1993). *Causation, Prediction, and Search*. Springer-Verlag.

Whittaker, J. (1990). *Graphical Models in applied multivariate statistics*. Wiley.

## 5 APPENDIX A — PROOF SKETCHES

The details given in this section are a bare-bones sketch of the proofs of the theorems in this paper. Detailed proofs can be found in the following section. Let $G$ be some directed acyclic graph and $\pi_G$ be the set of linearly independent parameters needed to encode any multinomial or multivariate normal distribution for which $G$ is an I-map. As the context demands, I let $\pi_G$ represent the parameters for either a multinomial or multivariate normal distribution.

**Claim 1** *The independence facts not entailed by d-separation applied to directed acyclic graph $G$ hold only for values of the parameters which satisfy non-trivial polynomial constraints.*

The proof of this claim is in two parts. First one can show, based upon the specific parameterization (multinomial or multivariate normal) that the constraints are polynomials in the parameters. Second I show that the constraints are non-trivial (not all value of the parameters satisfy the constraints). The proof of the non-triviality is similar to the main lemma used in the atomic-completeness result of Geiger et al..

**Claim 2** *For independence statement i not entailed by $G$ and for the Lebesgue measure over the set of parameters $\pi_G$ the set of values where the independence fact i holds is Lebesgue measure zero.*

The proof of this claim follows from the fact that the solution set to non-trivial polynomial constraints has measure zero (See Okamoto 1973).

Theorem 7 and Theorem 8 follow from Claim 2. With respect to a given graph $G$, only a finite number of independence facts are not faithfully entailed. Each of these independence facts hold only for a set of parameterizations of measure zero. The union of all of these finitely many sets of parameterizations is measurable and is of Lebesgue measure zero.

Theorem 5 and Theorem 6 follow from Theorem 7 and Theorem 8 by the following measure-theoretic argument. Given that the set of parameterizations in which the distribution is unfaithful are of measure zero and that there are sets of (permissible) parameterizations with positive measure then there are parameterizations which are faithful.

Finally, Theorem 3 and Theorem 4 and the strong completeness for $S$ sentences with respect to $\vdash_{\mathcal{F}}$ follow from the existence of faithful distributions for the two classes of distributions and the soundness of d-separation (Pearl 1988). All and only the independence facts which follow from the rule of d-separation hold in the faithful distribution. Theorem 1 and Theorem 2 are trivial consequences of Theorem 3 and Theorem 4.

# 6   APPENDIX B — PROOFS

A discrete Bayesian network is a tuple $\langle G, P \rangle$ where $P$ is a probability density function (with respect to the counting measure) over a finite set of variables $\mathbf{V}$ (each of which take on at least 2 values) and $G$ is a graph over the same set of variables $\mathbf{V}$ such that there exists a factorization of $P(\mathbf{V})$ such that

$$P(\mathbf{V}) = \prod_{A \in \mathbf{V}} P(A|parents(A))$$

where $P(A|parents(A))$ is a conditional probability density and $parents(A)$ is the set of parents of vertex $A$ in graph $G$. Linear Bayesian networks are described in Spirtes et al. (1993).

## 6.1   Parameterizations of Bayesian networks

Given that the joint distribution $P$ factors according to the graph $G$ into conditional densities one can parameterize the joint distribution by parameterizing each of the conditional distribution. (Only the cases where the the density is a joint density for a discrete Bayesian network and where the joint density is multivariate normal.) For variable $A \in \mathbf{V}$ define NV($A$) as the number of possible values that $A$ can take and C($A$) as the set of possible values of $A$. Let $inst(\emptyset) = 1$ and $inst(\{A1, ..., An\}) = NV(A1) \times ... \times NV(An)$. For each conditional distribution $P(A|parents(A))$ one can represent the conditional distributtion with $nparam(A, parents(A))$ linearly independent parameters where

$nparam(A, parents(A)) = (\text{NV}-1) \times inst(parents(A))$.

The reason for the (NV($A$) - 1) is that for any given instantiation of the parents the NV($A$)$^{th}$ probability is a linear combination of the other (NV($A$) - 1) parameters. I adopt the following convention for naming the parameters. $\theta_{D,d,(a1,...,aN)}$ is the parameter for the conditional probability $P(D = d|A1 = a1, ..., AN = aN)$ where $parents(D) = A1, ..., AN$ and where the variables $\langle A1, ..., AN \rangle$ are ordered lexicographically. Let $\theta$ be the set of all of the parameters for all of the variables in $\mathbf{V}$. Each of the parameters $\theta_{D,d,(a1,...,aN)}$ satisfies the constraints that $0 \le \theta_{D,d,(a1,...,aN)} \le 1$ and for all $a1, ..., aN$ it is the case that $\sum_{d \in C(D)} \theta_{D,d,(a1,...,aN)} = 1$.[8] The parameterization of linear Bayesian networks is discussed in Spirtes et al. (1993).

## 6.2   Faithfulness

Let $P$ be a probability density function over $\mathbf{V}$ and $G$ be a graph over the vertices in $\mathbf{V}$. $\langle G, P \rangle$ are said to satisfy the Markov condition if and only if for all $A \in \mathbf{V}$ it is the case that $A \perp\!\!\!\perp \mathbf{V} \backslash (parents(A) \cup$

---

[8]Note that if the distribution P is not positive then some of the parameters are not strictly necessary to parameterize the distribution.

---

$descendants(A))$ given the $parents(A)$. If $P$ is a (discrete) probability density function, i.e. a density with respect to the counting measure, then $\langle G, P \rangle$ is a discrete Bayesian network if and only if $\langle G, P \rangle$ satisfies the Markov condition. If $P$ is a multivariate normal density function then $\langle G, P \rangle$ is a Gaussian Bayesian network if and only if $\langle G, P \rangle$ satisfies the Markov condition. A conditional independence relation is entailed for graph G if and only if it is true in all distributions $P$ such that $\langle G, P \rangle$ satisfies the Markov condition. Let $\langle G, P \rangle$ be a discrete (Gaussian) Bayesian network. A discrete (Gaussian) Bayesian network is faithful if and only if the conditional independence relations true in $P$ are exactly those entailed by the factorization of $P$ with respect to $G$.

## 6.3   Method for constructing constraints

In this section I give a method for calculating the polynomial constraint that must be satisfied for a violation of faithfulness to occur.

Let $\langle G, P \rangle$ satisfy the Markov condition and let $A \perp\!\!\!\perp B | \mathbf{C}$ be an independence fact true in $P$ but such that $A \perp\!\!\!\perp B | \mathbf{C}$ is not implied by the Markov condition applied to $G$.

$$A \perp\!\!\!\perp B | \mathbf{C} \Leftrightarrow$$

$\forall (a, b, \mathbf{c})$ if $P(\mathbf{C} = \mathbf{c}) \ne 0$ then
$P(A = a, B = b | \mathbf{C} = \mathbf{c}) =$
$P(A = a|\mathbf{C} = \mathbf{c})P(B = b|\mathbf{C} = \mathbf{c}) \Leftrightarrow$    (*)

$\forall (a, b, \mathbf{c})$ if $P(\mathbf{C} = \mathbf{c}) \ne 0$ then
$P(A = a, B = b, \mathbf{C} = \mathbf{c})P(\mathbf{C} = \mathbf{c}) =$
$P(A = a, \mathbf{C} = \mathbf{c})P(B = b, \mathbf{C} = \mathbf{c})$

Thus, for a violation of faithfulness to occur a set of $inst(\{A, B\} \cup \mathbf{C})$ many equations must be satisfied. Now I will show that each of these equations is a polynomial in the parameters of the model. Let $\Omega = Ancestor(A, B, \mathbf{C})$.

$$P(\Omega) = \sum_{\mathbf{V} \backslash \Omega} P(\mathbf{V}) = \prod_{X \in \Omega} P(X|parents(X)) \quad (1)$$

Notice that all of the probabilities which occur in the last statement in (*) are of the form $P(\mathbf{F}=\mathbf{f})$ for some set of vertices $\mathbf{F}$ and some instantiations of those vertices $\mathbf{f}$. For instance if $\mathbf{F} = \{A\} \cup \{B\} \cup \mathbf{C}$ then $P(A=a, B=b, \mathbf{C}=\mathbf{c})$ is of the form $P(\mathbf{F}=\mathbf{f})$.

$$P(\mathbf{F} = \mathbf{f}) = \sum_{\Omega \backslash \mathbf{F}} P(\mathbf{F} = \mathbf{f}, \Omega \backslash \mathbf{F}) \quad (2)$$
$$= \sum S(i) \quad (3)$$

where each summand $S(i)$ is a product of $|\Omega|$ many parameters (see equation 1). There are $I = inst(\Omega \backslash \mathbf{F})$ many summands. Let $O1, ..., OI$ be the $I$ instantiations of the variables in $\Omega \backslash \mathbf{F}$. Define $M(A, i)$ as follows

$M(A, i) = $ the value of $A$ in the $i^{th}$ instantiation of $\Omega \backslash \mathbf{F}$ if $A \in \Omega \backslash \mathbf{F}$

= the value of $A$ in the instantiation $\mathbf{f}$ if $A \in \mathbf{F}$.

The $i^{th}$ summand $S(i)$ for (2) is formed in the following fashion. Let $p(A, j)$ be the $j^{th}$ parent of $A$ with respect to the lexicographic ordering over the set of parents of $A$.

$$S(i) = \prod_{A \in \Omega} \theta_{A,M(A,i),(M(p(A,1),i),\ldots,M(p(A,n),i))}$$

where $A$ has $n$ parents

Since the constraints are products of terms of the form $P(\mathbf{F}=\mathbf{f})$ given in (2) the constraints are polynomials in the parameters of the discrete Bayesian network.

Violations of faithfulness occur in the linear case for distributions only if polynomial constraints in the parameters of the model hold. This is shown in Spirtes et al. (1993).

### 6.4 The polynomial constraints are non-trivial

A polynomial in $n$ variables is said to be non-trivial (not an identity) if not all instantiations of the n variables are solutions of the polynomial. Now I show that all of the polynomials for non-entailed independence constraints are non-trivial.

This is done by using the property weak transitivity which is guaranteed to hold in Gaussian and discrete distributions where all of the variables are binary (see Pearl 1988). Weak transitivity allows us to give an alternative proof of the completeness of d-separation and a measure theoretic result about faithfulness for the Gaussian case as well as the discrete case.

Some inference rules about independence and dependence for probability theory (see Dawid 1979 and Pearl 1988).

$\neg X \perp\!\!\!\perp Y | Z \Rightarrow \neg Y \perp\!\!\!\perp X | Z$    (Symmetry)

$\neg X \perp\!\!\!\perp Y | Z \Rightarrow \neg X \perp\!\!\!\perp WY | Z$    (Decomposition)

For positive distributions
$\neg X \perp\!\!\!\perp WY | Z \wedge X \perp\!\!\!\perp W | ZY \Rightarrow \neg X \perp\!\!\!\perp Y | ZW$    (Intersection)

The following rule also holds for Gaussian and Boolean systems
$\neg X \perp\!\!\!\perp \gamma | Z \wedge \neg \gamma \perp\!\!\!\perp Y | Z \Rightarrow \neg X \perp\!\!\!\perp Y | Z \vee \neg X \perp\!\!\!\perp Y | Z\gamma$
(Weak Transitivity)

where $\gamma$ is a singleton set.

**Lemma 9 (Geiger et al.)** *If in directed acyclic graph $G$ there exists a d-connecting path between $A$ and $B$ given $\mathbf{C}$ then there exists a singly-connected subgraph $G'$ of $G$ such that $A$ and $B$ given $\mathbf{C}$ are d-connected by a path $p$ in $G'$ and such that the only edges in $G'$ are edges on the d-connecting path and a set of edges which form exactly one directed path from each collider on the path $p$ to a member of $\mathbf{C}$.*

**Proof** — Let $p$ be a d-connecting path between $A$

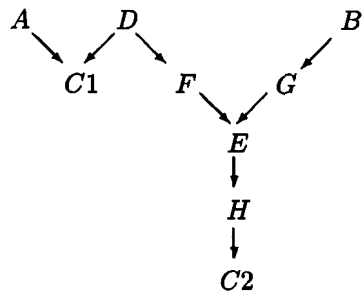

Figure 1: Schematic of singly connected graph between $A$ and $B$ given $\mathbf{C} = \{C1, C2\}$.

and $B$ given $\mathbf{C}$ in $G$. Let $G1$ be the subgraph of $G$ such that all of the edges on $p$ are in $G_1$ and for each collider $D$ on $p$ not in $\mathbf{C}$ include the edges that are on one path from $D$ to a member of $\mathbf{C}$ not through another member of $\mathbf{C}$. Arbitrarily choose one path if there are more than one from $D$ to members of $\mathbf{C}$.

Let $r(G_1)$ be the number of multiple pathways that exist in G1. It is clearly finite. If $r(G_1) > 0$ then there exists two distinct colliders on $p$, the d-connecting path from $A$ to $B$ given $\mathbf{C}$ in $G_1$, which have the same member of $\mathbf{C}$ as a descendent. Let $D1$ and $D2$ be such colliders. Let $p[D1, D2]$ be the set of edges between $D1$ and $D2$ on the path $p$. Remove $p[D1, D2]$ from the graph $G1$. Clearly $r(G_1)$ is reduced by removing $p[D1, D2]$. Continue the process until $r(G_1) = 0$. Clearly a d-connecting path between $A$ and $B$ given $\mathbf{C}$ remains at each stage.

The graph $G_1$ is the desired graph $G'$. □

Schematically, the claim amounts to the claim that there is a subgraph which looks like the graph in Figure 1 where $\mathbf{C} = \{C1, C2\}$.

Let $P$ be a probability distribution and $G$ be a directed acyclic graph. $\langle G, P \rangle$ satisfies the *local dependence* condition if and only if $\langle G, P \rangle$ is a Bayesian network and if $A \rightarrow B$ is in $G$ then $\neg A \perp\!\!\!\perp B$ is true in $P$.

**Claim 3** *For a singly-connected graph $G$ there exists a positive binary probability distribution $P$ distribution such that $\langle G, P \rangle$ satisfy the local dependence condition.*

**Claim 4** *For a singly-connected graph $G$ there exists a positive Gaussian probability distribution $P$ distribution such that $\langle G, P \rangle$ satisfy the local dependence condition.*

Both Claim 3 and Claim 4 are easy to show.

**Lemma 10** *If $\langle G, P \rangle$ satisfies the local dependence condition, $P$ is weakly transitive, $G$ is singly-connected, and there exists a directed path from $A_1$ to $A_n$ in $G$ then $\neg A_1 \perp\!\!\!\perp A_n$ is true in $P$.*

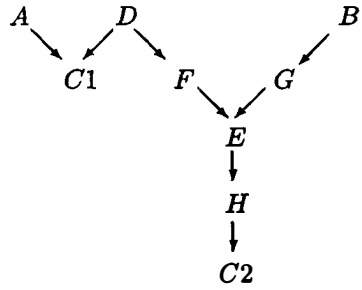

Figure 2: Graph $G'$.

**Proof** — by induction on length of path using weak transitivity.
*base case* — consider the trivial case of the null path from $A_1$ to $A_1$.
*induction step* — Assume that there is a path from $A_1$ to $A_n$ in $G$ and $\neg A_1 \perp\!\!\!\perp A_{n-1}$. $\neg A_{n-1} \perp\!\!\!\perp A_n$ follows from local dependence and $A_1 \perp\!\!\!\perp A_n | A_{n-1}$ by the Markov condition and the single-connectedness of $G$. Then, by weak transitivity, $\neg A_1 \perp\!\!\!\perp A_n$. □

The following lemma is the main step to proving Geiger et al.'s atomic completeness result. The following is an alternative proof which handles both the discrete and Gaussian cases simultaneously.

**Lemma 11** *For any directed acyclic graph $G$ which does not entail $A \perp\!\!\!\perp B | \mathbf{C}$ there exists a discrete binary (Gaussian) distribution $P$ such that $\neg A \perp\!\!\!\perp B | \mathbf{C}$ is true in $P$ and $\langle G, P \rangle$ is a discrete (Gaussian) Bayesian network.*

**Proof** — assume that $A \perp\!\!\!\perp B | \mathbf{C}$ is not entailed by Markov condition applied to $G$. Construct a binary valued (Gaussian) distribution $P_1$ over the variables in $G$ such that $\langle G, P \rangle$ is a discrete Bayes network and such that $\neg A \perp\!\!\!\perp B | \mathbf{C}$ is true in $P_1$. Since $A \perp\!\!\!\perp B | \mathbf{C}$ is not entailed by Markov condition applied to $G$ there must exist a path which d-connects $A$ and $B$ given $\mathbf{C}$. Let $G'$ be the subgraph of $G$ described in Lemma 9. To simplify the proof I will give an informal argument which can readily be turned into a rigorous inductive argument. Let $G'$ be described by the graph in Figure 2 where $\mathbf{C} = \{C1, C2\}$.

Let $P$ be a positive binary (Gaussian) probability distribution such that $\langle G', P \rangle$ satisfies the local dependence condition; one exists by Claim 3 (Claim 4). Note that the positivity of $P$ allows us to use Intersection as a rule of inference.

The goal is to show that $\neg A \perp\!\!\!\perp B | C1, C2$

(1) $\neg A \perp\!\!\!\perp D | C1$
proof —
(i) $A \perp\!\!\!\perp D$      Markov condition (applied to $G'$)
(ii) $\neg D \perp\!\!\!\perp C1 | \emptyset$      local dependence
(iii) $\neg C1 \perp\!\!\!\perp A | \emptyset$      local dependence

(iv) $\neg A \perp\!\!\!\perp D | C1$    weak trans. (WT), (i), (ii) and (iii)

(2) $\neg A \perp\!\!\!\perp D | C1, C2$
proof —
(i) $\neg A \perp\!\!\!\perp D, C2 | C1$      from Decomposition and (1)
(ii) $A \perp\!\!\!\perp C2 | D, C1$      from Markov condition
(iii) $\neg A \perp\!\!\!\perp D | C1, C2$      from (i), (ii) and Intersection

(3) $\neg D \perp\!\!\!\perp B | C2$
proof —
(i) $D \perp\!\!\!\perp B | \emptyset$      from Markov condition
(ii) $\neg D \perp\!\!\!\perp C2 | \emptyset$      local dependence and Lemma 10
(iii) $\neg C2 \perp\!\!\!\perp B | \emptyset$      local dependence and Lemma 10
(iv) $\neg D \perp\!\!\!\perp B | \emptyset$ or $\neg D \perp\!\!\!\perp B | C2$ from WT, (ii) and (iii)
(v) $\neg D \perp\!\!\!\perp B | C2$      from (i) and (iv)

(4) $\neg D \perp\!\!\!\perp B | C1, C2$
proof- as in proof of (2).

(5) $\neg A \perp\!\!\!\perp B | C1, C2$
proof —
(i) $A \perp\!\!\!\perp B | C1, C2, D$      from Markov condition
(ii) $\neg A \perp\!\!\!\perp B | C1, C2$ or
     $\neg A \perp\!\!\!\perp B | C1, C2, D$      from WT, (4) and (2)
(iii) $\neg A \perp\!\!\!\perp B | C1, C2$      from (i) and (ii)

Thus it has been established that $\neg A \perp\!\!\!\perp B | C1, C2$ and it is clear that it is possible to extend $P_1$ to a distribution $P$ over $G$. Let $\mathbf{V}'$ be the set of vertices in $G'$ and $\mathbf{V}$ be the set of vertices in $G$ and let $\{Z_1, Z_2, ..Z_n\}$ be an enumeration of the vertices in $\mathbf{V} \setminus \mathbf{V}'$. In the discrete case let $P(\mathbf{V}) = P_1(\mathbf{V}') P(Z_1) \ldots P(Z_n)$ where $P(Z_i)$ is any arbitrary binary distribution over the variable $Z_i$. In the Gaussian case let $\text{cov}(Z_i, X) = 0$ for all $0 \leq i \leq n$ and $X \in \mathbf{V}'$ and set the variances of $Z_i$ arbitrarily but not to zero. It should be clear that the proof above can be turned into an induction over the number of directed paths (or more exactly semi-treks) in the d-connecting path in $G'$. □

**Lemma 12 (Atomic completeness; Geiger et al.)** *For any directed acyclic graph $G$ over variables $\mathbf{V}$ which does not entail $A \perp\!\!\!\perp B | \mathbf{C}$ there exists a discrete (not necessarily binary) distribution $P$ such that $\neg A \perp\!\!\!\perp B | \mathbf{C}$ is true in $P$ and $\langle G, P \rangle$ is a discrete Bayesian network.*

**Proof** — Begin by constructing the discrete binary distribution $P$ from Lemma 11. Simply expand the distribution based on binary valued probabilities to one based upon the number of categories required for each of the variables in $\mathbf{V}$; the resulting distribution is essentially a binary distribution extended to an arbitrary discrete probability space by using zero probabilities. Assume that the values of the binary variables in $P$ are either zero or one (0 or 1). The easiest way to extend the distribution is to force the probability of $\mathbf{V} = \mathbf{v}$ to be zero if for some $A \in \mathbf{B}$ the value of $A$ in $\mathbf{v}$ is not either 0 or 1. The dependence follows since all of the polynomials described in equation (*) must hold for the independence to hold and by Lemma 11 and this is not the case. □

**Theorem 1 (Geiger et al.)** $G \models_{\mathcal{P}_D} i$ if and only if $G \vdash_{D_*} i$.

**Proof** — This follows from Lemma 12 and the soundness of d-separation (see Pearl 1988).□

**Theorem 2 (Geiger et al.)** $G \models_{\mathcal{P}_N} i$ if and only if $G \vdash_{D_*} i$.

**Proof** — This follows from Lemma 11 and the soundness of d-separation.□

### 6.5 Polynomial constraints are Lebesgue measure zero and the completeness of d-separation.

**Lemma 13 (Okamoto)** - *The solutions to a (nontrivial) polynomial are Lebesgue measure zero over the space of the parameters of the polynomial.*

For a fixed statespace (i.e. the number of categories for each variable) let $\pi_G^{\mathcal{D}}$ be the set of linearly independent parameters needed to parameterize an arbitrary discrete distribution for which graph $G$ is an I-map and let $\pi_G^{\mathcal{N}}$ be the set of linearly independent parameters needed to parameterize an arbitrary multivariate normal distribution for which graph $G$ is an I-map. For the discrete case, the set of legal parameterizations $E \subseteq [0,1]^n$ where $n$ is the number of linearly independent parameters. For the Gaussian case, the set of legal parameterizations is the space $\mathcal{R}^n$.

**Lemma 14** *For a fixed statespace with n linearly independent parameters, the set of parameterizations $\omega$ over a graph $G$ in which independence fact $A \perp\!\!\!\perp B | \mathbf{C}$ is true but such that $A \perp\!\!\!\perp B | \mathbf{C}$ has measure zero with respect the Lebesgue measure over $\mathcal{R}^n$.*

**Proof** — Let $n = inst(\{A, B\} \cup \mathbf{C})$. There are $n$ polynomials which must hold for this violation to occur. The polynomials are non-trivial by Theorem 11. Let $\omega_i$ be the set of solutions to the $i^{\text{th}}$ polynomial.

$$\omega = \bigcap_{i=1}^{n} \omega_i$$

$\omega$ is measurable since finite intersections of measurable sets are measurable. Let $\omega' = \bigcup_{i=1}^n \omega_i$. Since $\omega'$ is the finite union of measurable sets it is measurable. $\mu(\omega) \leq \mu(\omega') \leq \sum_{i=1}^n \mu(\omega_i) = 0$ and given the non-negativity of the measure $\mu(\omega) = 0$; $\Omega$ is Lebesgue measure zero.□

**Theorem 7 (Measure zero)** *With respect to the Lebesgue measure over $\pi_G^{\mathcal{D}}$, the set of distributions which are unfaithful to $G$ is measure zero.*

**Proof** — There are a finite number of sets of polynomials which must be satisfied to violate faithfulness. Let $n$ be the number of such polynomials and $\omega_i$ be the set of solutions to the $i^{\text{th}}$ set of polynomials, each of these sets is of measure zero by Lemma 14.

$$\omega = \bigcup_{i=1}^{n} \omega_i$$

$\omega$ is measurable since finite unions of measurable sets are measurable and the set $\omega$ is also Lebesgue measure zero. Finally restrict the solution set $\omega$ to the interval [0,1]. Let $E \subseteq [0,1]^n$ be the subset of legal parameterizations of a distribution where $n$ is the dimensionality of the space for the Lebesgue measure, $\mu$. E is a closed set and thus measurable. As E is a measurable set and the finite intersection of measurable sets is again measurable, $\omega \cap E$ is measurable. Since $\omega \cap E \subset \omega$ it is the case that $\mu(\omega \cap E) \leq \mu(\omega) = 0$ and by the non-negativity of the measure $\mu$ it is the case that $\mu(\omega \cap E) = 0$.□

**Theorem 8 (Measure zero; Spirtes et al.)** *With respect to the Lebesgue measure over $\pi_G^{\mathcal{N}}$, the set of distributions which are unfaithful to $G$ is measure zero; i.e. Violations of faithfulness in Gaussian probability distributions are Lebesgue measure zero.*

**Proof** — Similar to proof of Theorem 7.□

**Theorem 5** *For all directed acyclic graphs $G$ there exists a $P \in \mathcal{P}_D$ which is faithful to $G$.*

**Proof** — Follows from Theorem 7 by the following measure-theoretic argument. Given that the set of parameterizations in which the distribution is unfaithful are of measure zero and that there are sets of (permissible) parameterizations with positive measure then there are parameterizations which are faithful.□

**Theorem 6 (Geiger et al. Spirtes et al.)** *For all directed acyclic graphs $G$ there exists a $P \in \mathcal{P}_N$ which is faithful to $G$.*

**Proof** — Follows from Theorem 8 as in Theorem 5.□

**Theorem 3 (Strong completeness)** $G \models_{\mathcal{P}_D} I$ *if and only if $G \vdash_{D_*} I$; i.e. d-separation is strongly complete for the class of multinomial distributions over arbitrary directed acyclic graphs.*

**Proof** — This follows from the existence of a faithful distribution in the multinomial class of distributions (Theorem 5) and the soundness of d-separation.□

**Theorem 4 (Geiger et al., Spirtes et al.)** $G \models_{\mathcal{P}_N} I$ *if and only if $G \vdash_{D_*} I$; i.e. d-separation is strongly complete for the class of Gaussian probability distributions over arbitrary directed acyclic graphs.*

**Proof** — This follows from the existence of a faithful distribution in the class of Gaussian distributions (Theorem 6) and the soundness of d-separation.□